\documentclass{article}

\usepackage{arxiv}

\usepackage[utf8]{inputenc} 
\usepackage[T1]{fontenc}    
\usepackage{hyperref}       
\usepackage{url}            
\usepackage{booktabs}       
\usepackage{amsfonts}       
\usepackage{nicefrac}       
\usepackage{microtype}      
\usepackage{lipsum}
\usepackage{moreverb,url}
\usepackage{graphicx}
\usepackage{amssymb,amsmath}
\usepackage{subcaption}
\usepackage{comment}
\usepackage{xcolor}
\usepackage{comment}
\usepackage{multirow}
\usepackage{relsize}

\usepackage{fouriernc} 

\newcommand{\PB}[1]{\textcolor{red}{PB: #1}}

\title{\LARGE Deep-Ensemble-Based Uncertainty Quantification in \\ Spatiotemporal Graph Neural Networks for Traffic Forecasting
}

\author{
  Tanwi Mallick \\
  Mathematics and Computer Science Division\\
  Argonne National Laboratory, Lemont, IL \\
  \texttt{tmallick@anl.gov} \\
   \And
 Prasanna Balaprakash \\
  Mathematics and Computer Science Division \& 
  Argonne Leadership Computing Facility \\
  Argonne National Laboratory, Lemont, IL \\
  \texttt{pbalapra@anl.gov} \\
  \And
  Jane Macfarlane\\
  Sustainable Energy Systems Group \\
  Lawrence Berkeley National Laboratory, Berkeley, CA \\
  \texttt{jfmacfarlane@lbl.gov} \\
}
\begin{document}

\maketitle
\thispagestyle{plain}
\pagestyle{plain}

\begin{abstract}
Deep-learning-based data-driven forecasting methods have produced impressive results for traffic forecasting. A major limitation of these methods, however, is that they provide forecasts without estimates of uncertainty, which are critical for real-time deployments. We focus on a diffusion convolutional recurrent neural network (DCRNN), a state-of-the-art method for short-term traffic forecasting. We develop a scalable deep ensemble approach to quantify uncertainties for DCRNN. Our approach uses a scalable Bayesian optimization method to perform hyperparameter optimization, selects a set of high-performing configurations, fits a generative model to capture the joint distributions of the hyperparameter configurations, and trains an ensemble of models by sampling a new set of hyperparameter configurations from the generative model. We demonstrate the efficacy of the proposed methods by comparing them with other uncertainty estimation techniques. We show that our generic and scalable approach outperforms the current state-of-the-art Bayesian  and number of other commonly used frequentist techniques.

\end{abstract}


\section{Introduction}


Traffic forecasting is a foundational component of an intelligent transportation
system. Precise forecasting across normal and extreme traffic conditions is crucial to improving traffic control, mitigating congestion, and resolving other traffic-related issues. Recently, data-driven deep learning approaches \cite{yu2017spatio, li2017diffusion, zheng2019gman} have received increased attention in traffic forecasting because of their impressive accuracy. 
However, uncertainty quantification (UQ) for traffic forecasting has received limited attention from the research community.  UQ is essential for understanding the limitations of the predictive models and how to use the models in an active traffic management system. 

Uncertainty in traffic forecasting can be decomposed into two categories: aleatoric uncertainty, which is irreducible uncertainty in the data, and epistemic uncertainty, which is uncertainty in the model \cite{tagasovska2018single}.  Aleatoric uncertainty occurs because of the inherent variability in traffic data. For example, at a given time of day, the flow and speed vary greatly. Or, faulty sensors may cause aleatoric uncertainty. This form of uncertainty is a data property, not a predictive model attribute. With aleatoric uncertainty, gathering more data in the same manner or employing a better model cannot reduce it. In contrast, regions that are underrepresented in the training dataset (out-of-distribution generalization \cite{lakshminarayanan2016simple}) can cause a mismatch between model estimation and data distribution. This situation creates epistemic uncertainty. In this case, epistemic uncertainty may be reduced by collecting more data.
Proper estimation of both the aleatoric and epistemic uncertainty for traffic forecasting is critical for understanding the limitations of the model. Furthermore, determining these uncertainty measures can be useful for detecting sensor locations where the data is most uncertain and taking measures to improve the data collection. 

For decades, traffic forecasting techniques have been extensively investigated using integrated autoregressive moving average \cite{williams2003modeling}, Kalman filters \cite{kumar2017traffic}, support vector machines \cite{castro2009online, ahn2016highway}, and artificial neural networks \cite{chan2012neural,karlaftis2011statistical}. Recently, deep learning techniques have received attention because of the impressive forecasting accuracy compared with statistical approaches and traditional machine learning approaches. Researchers have explored recurrent neural networks (RNNs) and long short-term memory  networks \cite{ma2015long} and convolutional neural networks  \cite{zhang2016dnn, zhang2017deep} 
for short-term traffic forecasting. Recently, graph neural networks such as spatiotemporal graph convolutional networks \cite{yu2017spatio}, diffusion convolutional recurrent neural networks (DCRNN) \cite{li2017diffusion}, and graph multi-attention networks \cite{zheng2019gman} have achieved state-of-the-art performance by capturing the spatiotemporal dynamics of the road network. 
However, without proper quantification of the uncertainty, forecast accuracy is not trustworthy for using the model in practice. 

By estimating the aleatoric and epistemic uncertainty separately for traffic forecasting models, insight into whether the predictive uncertainty comes from the model or is inherently present in the data can provide insight into how to best use the model.
For example,  the 90\% quantile or 10\% quantile of the aleatoric uncertainty measure provide will predictions for the worst and best traffic conditions in a given location. 
This information can then be used to inform rerouting algorithms and other proactive traffic management strategies. 
On the other hand, if for example, the structure of the road network changes, the distribution of the training and test data may no longer represent the associated changes in the traffic patterns. This will increase the model’s epistemic uncertainty and capture the model’s lack of confidence in the forecasting results (out-of-distribution generalization \cite{lakshminarayanan2016simple}). In the absence of additional data for training, the epistemic uncertainty can be used to establish trustworthiness and to know when and where the model predictions are reliable.  


We focus on DCRNN \cite{li2017diffusion}, one of the most promising approaches to traffic forecasting. 
It has achieved state-of-the-art performance by capturing both spatial and temporal dependencies of the traffic network through a combination of graph and recurrent neural networks. Our proposed approach for uncertainty estimation in DCRNN is comprised of the following steps:
\begin{itemize}
    \item The DCRNN is configured to model the distribution of traffic data using simultaneous quantile regression (SQR).
    \item A scalable Bayesian optimization method is used to tune the hyperparameters of the DCRNN model with simultaneous quantile regression.
    \item A set of high-performing hyperparameter configurations from the hyperparameter optimization is used to fit a generative model; this captures the joint probability distribution of the high-performing hyperparameter values. A new set of hyperparameter configurations is then sampled from the generative model, and the corresponding DCRNN model are trained simultaneously.
    \item A set of high performing models is selected for ensemble construction and a variance decomposition approach is then used to estimate the aleatoric and epistemic uncertainties from the ensemble predictions.   
\end{itemize}
The main contributions of the paper are:
\begin{itemize}
    \item A scalable approach to construct deep ensembles using Bayesian hyperparameter optimization and generative modeling,  
    \item A conceptually simple, yet effective, Gaussian assumption-free uncertainty quantification approach to quantify uncertainty of a DCRNN model,
    \item The first demonstration of aleatoric and epistemic uncertainty estimation for short-term traffic forecasting, and
    \item A demonstrated improvement over the state-of-the-art Bayesian DCRNN trained by stochastic gradient Markov Chain Monte Carlo for uncertainty estimation in DCRNN.
\end{itemize}

\section{Diffusion Convolutional Recurrent Neural Network (DCRNN)}
The DCRNN considers the road network as a weighted directed graph $G = (V, E, A)$, where the set $V$ of nodes represents $N$ traffic measurement (sensor) locations, the set $E$ of edges represents the connectivity between the nodes,
and the adjacency matrix  $A$ ($A \in R^{N\times N}$) represents the proximity between the nodes. 
Given the graph $G$ and historical time series data of traffic metrics $X$,  DCRNN seeks to learn a function \text{f(.)} to forecast short-term traffic metrics for $K$ time steps given the past $K'$ traffic metrics as input:
\begin{equation*}
X(k-K'+1), ...,X(k); G \xrightarrow{\text{f(.)}} X(k+1),... , X(k+K),
\end{equation*}
where $X(k) \in \mathcal{R}^{N \times 1}$ is a vector of traffic metrics defined over $N$ vertices at time step $k$. 


 DCRNN is an encoder-decoder neural network architecture with RNN layers that operates on a graph. It models the temporal dependency of the traffic using gated recurrent units (GRUs), a variant of RNN. Specifically, the matrix multiplications of the GRU cell are replaced by a diffusion-convolutional operation
to model the spatial dependency of the road network. The modified GRU cell is defined as
\begin{equation}\label{equ_1}
\begin{array}{lcl}
r^k & = &  \sigma (W_{r\bigstar G} [X(k), h_{k-1}] + b_r) \\
u^k & = &  \sigma (W_{u\bigstar G} [X(k), h_{k-1}] + b_u) \\
c^k & = &  \tanh (W_{c\bigstar G} [X(k) (r_k \odot h_{k-1}] + b_c) \\
h_k & = & u_k \odot h_{k-1} + (1 - u_k) \odot c_k ,
\end{array}
\end{equation}
where $X(k)$ and $h_k$ represent the traffic state and final state
at time $k$, respectively;
$u_k$, $r_k$, and $c_k$ are the update gate, reset gate, and cell state, respectively; $\bigstar G$ denotes the diffusion-convolution operation; and $W_r, W_u,$ and $W_c$ are parameters of the GRU cell.
The diffusion-convolution operation $\bigstar G$ on the graph $G$ is defined as
\begin{equation}
\label{equ_2}
 W_{(\cdot)\bigstar G} X = \sum_{d=0}^{\mathcal{K}-1} (W_{(\cdot)O} (D_O^{-1}A)^d + W_{(\cdot){I}}(D_I^{-1}A)^d) X,
\end{equation}
where $\mathcal{K}$ is a maximum diffusion step;  $D_O$ and $D_I$ are the in-degree and out-degree diagonal matrices, respectively; $A$ is the adjacency matrix; and $W_O$ and $W_I$ are the learnable filters for the diffusion process and reverse one, respectively.

The encoder takes the input adjacency matrix and $K'$ time series data and maps the data into a latent representation. The decoder takes the latent representation and produces forecasts for future traffic for $K$ timestamps. In the training,  DCRNN optimizes the following mean absolute error (MAE) loss using a minibatch stochastic gradient algorithm:
\begin{align}
 \hat{f} \in \operatorname*{argmin}_{f}   \frac{1}{N} \mathlarger{\mathlarger{\sum}}_{j=1}^{N} \frac{1}{n} \mathlarger{\mathlarger{\sum}}_{i=1}^{n} | y_{ij} - \hat{y}_{ij}|, 
\label{mae}
\end{align}
where $y$ and $\hat{y}$ are, respectively, the observed and corresponding forecast values for the $i^{th}$ time step and $j^{th}$ sensor location;  $n$ is the total timestamps for each sensor locations; and $N$ is the total number of sensor locations.

\section{Methodology}

\begin{figure}[ht]
    \centering
    \includegraphics[width=\linewidth]{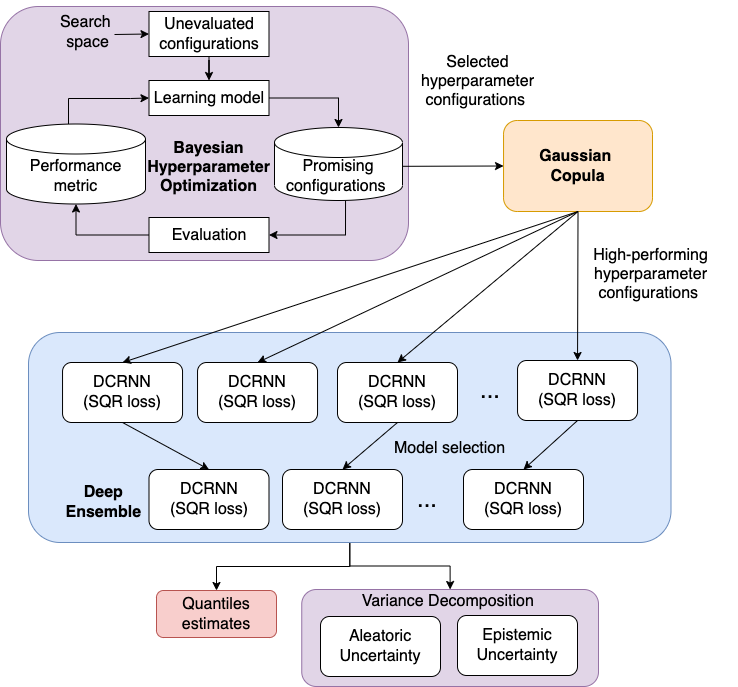}  
\caption{Components of our proposed approach DCRNN-SQRUQ: (1) simultaneous quantile regression (SQR) loss for DCRNN, (2) Bayesian hyperparameter optimization, (3)  high-performing hyperparameter generation using a Gaussian copula method, and (4) uncertainty estimation using variance decomposition for deep ensemble.}
\label{fig_diagram}
\end{figure}

We focus on uncertainty quantification in DCRNN predictions. Our proposed method, DCRNN with simultaneous quantile regression (DCRNN-SQRUQ),
seeks to  estimate both aleatoric (data) and epistemic (model) uncertainties. Our methodology has four steps, as illustrated in  Figure \ref{fig_diagram}: (1) simultaneous quantile regression (SQR) loss for  DCRNN, (2)  hyperparameter optimization using scalable Bayesian optimization method, (3) high-performing hyperparameter generation using a Gaussian copula model, and (4) aleatoric and epistemic uncertainty estimation using variance decomposition for deep ensemble. 

\subsection{Simultaneous quantile regression} \label{sqr}
The first step of our approach is the adaptation of SQR in  DCRNN to predict the quantiles of the forecasting traffic distribution.

Consider a typical regression setting consisting of input features $X$ and output variable $Y$ taking real values $y$. Let $F(y)=P(Y \leq y)$ be the cumulative distribution function of $Y$; the $q^{th}$ quantile of $Y$ can be calculated by the quantile distribution function $F^{-1}(q) = inf (y:F(y) \geq q)$, where $q \in (0, 1)$. The quantile regression approach seeks to model $\hat{y}=\hat{f}_{q}(x)$ that approximates the conditional quantile distribution function $y=F^{-1}(q \mid X=x)$. Specifically, the quantile regression estimates the value of $Y$ at a given quantile $q$ conditioned on the given $x$ values of input features $X$. This estimation can be achieved by using  the pinball loss function \cite{fox1964admissibility, koenker1978regression}:
\begin{equation}
\ell_{q}(y, \hat{y})=\left\{\begin{array}{lr}
q(y-\hat{y}) & \text { if } y-\hat{y} \geq 0 \\
(1-q)(\hat{y}-y) & \text { else }
\end{array}\right.
\end{equation}
Building on this, the recently proposed SQR approach estimates all the quantiles simultaneously by solving the following optimization problem:  
\begin{equation}
\hat{f} \in \arg \min \frac{1}{n} \sum_{i=1}^{n} \underset{q \sim U[0,1]}{\mathbb{E}}\left[\ell_{q}\left(f\left(x_{i}, q\right), y_{i}\right)\right] .
\label{sqr}
\end{equation}
This optimization problem is solved by using minibatch stochastic gradient descent, where random quantile levels $q \sim U[0,1]$ (uniform distribution) are sampled afresh for each training point and for each minibatch. 

In DCRNN, for each sensor location and for each forecast time step, we seek to estimate the traffic metric at all quantiles conditioned on the past $K'$ traffic metrics defined over $N$ sensor locations. This is achieved by replacing the MAE loss minimization with the SQR minimization. Applying Eq.~\ref{sqr} in Eq.~\ref{mae}, we get
\begin{equation}
    \hat{f} \in \operatorname*{argmin}_{f} \frac{1}{N} \mathlarger{\mathlarger{\sum}}_{j=1}^{N} \frac{1}{n} \mathlarger{\mathlarger{\sum}}_{i=1}^{n} \operatorname*{\mathbb{E}}_{q \sim U[0, 1]} [l_{q} (f(x_{ij}, q,G),y_{ij})].
\end{equation}

The key advantages of SQR include well-calibrated prediction intervals; joint estimation of the quantiles that can eliminate the unacceptable crossing quantiles \cite{takeuchi2006nonparametric}; and the ability to model non-Gaussian, skewed, asymmetric, multimodal, and heteroskedastic noise in the data.

\subsection{Scalable hyperparameter optimization}

The second step of our approach is the use of a scalable hyperparameter search for tuning the hyperparameters of DCRNN. 

DCRNN has several training hyperparameters, including  batch size, a threshold \texttt{max\_grad\_norm} to clip the gradient norm to avoid exploring gradient problem of RNN \cite{pascanu2013difficulty}, initial learning rate and learning rate decay, maximum diffusion steps, and architecture
configurations (network topology parameters) such as number of encoder and decoder layers,  number of RNN units per layers, and filter type (i.e., random walk, Laplacian). The hyperparameters are crucial in determining  DCRNN's forecasting accuracy.


To tune the hyperparameters of the DCRNN model with SQR, we use the DeepHyper  \cite{balaprakash2018deephyper} open source software that has a scalable Bayesian optimization (BO) method.  Given the training and validation data, a neural network model, and the feasible set of values of the training hyperparameters, DeepHyper uses an asynchronous BO method seeks to minimize the validation error by repeatedly training the neural network model with various hyperparameter values. The DeepHyper BO uses an incrementally updated random forest model to learn the relationship between the hyperparameter values and their corresponding validation errors. The BO method uses the random forest model to predict the validation errors of the evaluated hyperparameter configurations along with the uncertainty associated with the predicted value. The BO search  involves both evaluation of hyperparameter configurations, where the random forest model is most uncertain, and exploitation, focusing on hyperparameter configurations that are closer to the previously found high-performing configurations. The balance between  exploration and exploitation is achieved by selecting an unevaluated hyperparameter configuration $\boldsymbol{x}^{*}$ that minimizes the lower confidence bound ($\mathrm{LCB}$) acquisition function:
\begin{align}
\boldsymbol{x}^{*}=\underset{\boldsymbol{x} \in \mathcal{D}}{\arg \min } \underbrace{\hat{\mu}(\boldsymbol{x})-\lambda \times  \hat{\sigma}(\boldsymbol{x})}_{\mathrm{LCB}},
\end{align}
where $\hat{\mu}(\boldsymbol{x})$ and $\hat{\sigma}(\boldsymbol{x})$ are respectively the predicted estimate and the standard deviation of that estimate from the random forest model of the given unevaluated hyperparameter configuration $\boldsymbol{x}$ sampled from the hyperparameter space $\mathcal{D}$ and where $\lambda$ is a user-defined parameter that controls the trade-off between the exploration and exploitation. 


A key reason for choosing DeepHyper for the DCRNN hyperparameter tuning is scalability. The DeepHyper BO follows a single-manager multiple-workers parallization scheme, where the manager runs the search,  generates hyperparameter configurations, and sends them to workers for evaluation; each worker is responsible for evaluating a hyperparameter configuration by training the neural network model with the given configuration, computing the validation error, and returning that value to the manager. The manager generates hyperparameter configurations for evaluation in an asynchronous manner: as soon as one or more workers returns the validation error, the manager uses the returned values to retrain the surrogate model within the BO and generates the next hyperparameter configuration for the worker. To enable simultaneous multiworker evaluations, the BO uses a constant liar scheme, where configurations are generated sequentially by refitting the random forest model with a lie for the validation error, given by a predicted value from the model itself, for the previously selected unevaluated hyperparameter configuration.



We use DeepHyper to tune the hyperparameters of DCRNN not only to find the best hyperparameter configuration with minimal validation error but also to extract a set of hyperparameter values with minimal validation errors for generative modeling. Moreover, we scale the hyperparameter search on an increasing number of compute nodes to see how scaling impacts the overall uncertainty estimation in DCRNN. 

\subsection{Generative modeling and ensemble training}
The third step of our approach is generative modeling.  We model the joint probability distribution of the high-performing hyperparameter configurations from the previous step and use the distribution for  generating more high-performing hyperparameter configurations to train an ensemble of DCRNN models. 


Copula is a statistical approach used to model the joint probability distribution of a given set of random variables by analyzing the dependencies between their marginal distributions.
We use a multivariate Gaussian copula model \cite{schmidt2007coping} to learn the joint probability distribution between high-performing hyperparameter configurations.  
The copula model $C$ takes multiple marginal distributions of the hyperparameters  $u_1, u_2,..., u_{hp}$ and returns a multivariate distribution as follows:
\begin{equation}
    C(u_i, u_2, ..., u_{hp}) = \Phi_{\sum} (\Phi^{-1}(u_1),  \Phi^{-1}(u_2), ..., \Phi^{-1}(u_{hp})),
\end{equation}
where $\Phi_{\sum}$  represents the cumulative distribution function (CDF) of a multivariate normal, with covariance $\sum$ and mean 0, and $\Phi^{-1}$ is the inverse CDF for the standard normal. This captures the joint probability distribution of the high-performing hyperparameter values such as learning rate, batch size, number of layers, and objective function. 

The rationale for generative modeling through multivariate Gaussian copula is twofold. First, in hyperparameter search problems characterized by integer, real, and categorical hyperparameters, there are multiple high-performing regions; while BO cannot sample these regions uniformly, it can sample a few configurations from different high-performing regions. By selecting these high-performing configurations and fitting a Gaussian copula, we can sample from the sparsely sampled high-performing regions as well. Second, if we have to find more high-performing configurations from different regions directly with BO, then we have to run the search multiple times with different initialization; this process will be both computationally expensive and resource intensive.

Once the copula model $C$ is trained, we sample a number of hyperparameter configurations from  $C$ and run multiple DCRNN training sessions with the sampled hyperparameter configurations. From these trained DCRNN models, we select a set of high-performing models for ensemble construction. 


\subsection{Variance decomposition for DCRNN ensembles}

The fourth step of our approach is to use the  ensemble of DCRNN models to estimate the traffic metrics and the associated aleatoric and epistemic uncertainty using variance decomposition.


Let  $\mathcal{E} = {\theta_{i}, i = 1, 2,..., M}$ be the ensemble of $M$ models.  Previous studies have showed that if each model is configured to predict the mean and variance of an output distribution that is assumed to be  Gaussian and if the model is trained with negative log likelihood loss \cite{egele2021autodeuq},  the empirical estimation of the mean and the total variance can be written as follows.   
\begin{equation}
\begin{split}
\mu_{\mathcal{E}}     &  = \frac{1}{M}\sum_{\theta \in \mathcal{E} } \mu_{\theta} \\
\sigma_\mathcal{E}^2  &  = \underbrace{\frac{1}{M} \sum_{\theta \in \mathcal{E}} \sigma^2_\theta}_{ \text{Aleatoric Uncertainty} } + \underbrace{\frac{1}{M-1} \sum_{\theta \in \mathcal{E}} (\mu_\theta - \mu_\mathcal{E})^2}_{ \text{Epistemic Uncertainty}}
\label{eq:var-decomposition-1}
\end{split}
\end{equation}
The total uncertainty quantified by $\sigma_\mathcal{E}^2 $ is a combination of aleatoric and  epistemic uncertainty, which is given by the mean of the predictive variance of each model in the ensemble and the predictive variance of the mean of each model in the ensemble.  


Our proposed SQR-based DCRNN seeks to predict quantiles of the distribution as opposed to the mean and variance of the distribution. Thus it is more generic and free from Gaussian assumptions. We derive the empirical estimate of the median ($q_{50\%}$) and the total variance of the ensemble as
\begin{align}
\begin{aligned}
{q_{50\%}}_{\mathcal{E}}     &  = \frac{1}{M}\sum_{\theta \in \mathcal{E} } {q_{50\%}}_{\theta} \\
\sigma_\mathcal{E}^2  &  = \underbrace{\frac{1}{M} \sum_{\theta \in \mathcal{E}}  (\frac{{q_{66\%}}_\theta - {q_{33\%}}_\theta}{2})^2}_{ \text{Aleatoric Uncertainty} }\\  & \qquad \qquad \qquad +\\ 
 &  \qquad \underbrace{\frac{1}{M-1} \sum_{\theta \in \mathcal{E}} ( {q_{50\%}}_\theta - {q_{50\%}}_\mathcal{E})^2}_{ \text{Epistemic Uncertainty}},
\label{eq:var-decomposition-2}
\end{aligned}
\end{align}
where $q_{33\%}$ and $q_{66\%}$ are the 33\% and 66\% quantiles of the output distribution, respectively. 

Note that Eq. \ref{eq:var-decomposition-2} becomes Eq. \ref{eq:var-decomposition-1} when the output distribution is Gaussian, where the mean and median will be the  same, (${q_{50\%}}_{\theta} = \mu_\theta$), and 
$({q_{66\%}}_\theta - {q_{33\%}}_\theta)/2$ will be equal to $\sigma_\theta$.



\section{Experimental results}



In this section we extensively evaluate our proposed DCRNN-SQRUQ method and compare it with state-of-the-art UQ techniques proposed for DCRNN.
We conduct experiments on the METR-LA dataset, which contains speed profiles, given in miles per hour (mph), of 207 sensor locations collected from the loop detector on the highways of Los Angeles County. It contains time series data from March 1, 2012, to June 30, 2012, aggregated at 5-minute intervals, for a total of 34,272 timestamps. From the 4-month data, 70\%  (first 12 weeks approx.) is used for training, 10\% (1.7 weeks approx.) for validation, and 20\% (3.4 weeks approx.) for testing. 
This dataset is widely used for benchmarking traffic forecasting techniques \cite{li2018dcrnn_traffic, mallick2020graph, zheng2020gman} 
including the recent state-of-the-art UQ estimation work \cite{wu2021quantifying}. 
The default hyperparameter configuration used for DCRNN is as follows: batch size -- 64; filter type -- dual/bidirectional random walk (captures both the upstream and the downstream traffic dynamic);  number of diffusion steps -- 2; RNN layers -- 2; RNN units per layer-- 64; optimizer -- Adam;  threshold for gradient clipping -- 5; initial learning rate -- 0.01; and learning rate decay -- 0.1. The adjacency matrix of the graph connectivity has been built by using the driving distance between the sensor locations. The DCRNN model uses the past 60 minutes of time series data at each node of the graph to forecast traffic for the next 5, 10, 15, ..., 60 minutes. 
To evaluate the performance, we use the same metrics used in the current state-of-the-art DCRNN UQ estimation work \cite{wu2021quantifying}. These metrics are the  mean absolute error (MAE), mean interval score (MIS), and Interval score. 
MAE is defined as
\begin{equation}\label{eq_mae}
\text{MAE} =  \frac{1}{N} \mathlarger{\mathlarger{\sum}}_{j=1}^{N} \frac{1}{n} \mathlarger{\mathlarger{\sum}}_{i=1}^{n} | y_i - \hat{y}_i |,
\end{equation}
where $N$ is the total number of sensor locations and $n$ is the number of  timestamps for each sensor locations.
MIS and interval scores are defined as
\begin{equation}\label{eq_mis}
\begin{array}{cc}
    \text{MIS}(q_u,q_l,q) &= \frac{1}{N} \mathlarger{\mathlarger{\sum}}_{j=1}^{N} \frac{1}{n} \mathlarger{\mathlarger{\sum}}_{i=1}^{n}  (q_{u_{ij}}-q_{l_{ij}}) \\
    & + \frac{2}{\rho} (y_{ij} - q_{u_{ij}}) \mathbb{1} \{y_{ij} > q_{u_{ij}}\} \\
    & + \frac{2}{\rho} (q_{l_{ij}} - y_{ij}) \mathbb{1} \{y_{ij} < q_{l_{ij}}\}
    \end{array}
\end{equation}

\begin{equation}\label{eq_interval}
\text{Interval} = \frac{1}{N} \mathlarger{\mathlarger{\sum}}_{j=1}^{N} \frac{1}{n} \mathlarger{\mathlarger{\sum}}_{i=1}^{n}  (q_{u_{ij}}-q_{l_{ij}}),
\end{equation}
where  $N$ is the total number of sensor locations and $n$ is the number of  timestamps for each sensor locations.
The estimated upper and lower quantiles $q_u$ and $q_l$ are the $(1-\frac{\rho}{2})$ and $\frac{\rho}{2}$ quantile for the confidence interval $(1-\rho)$, and $y$ is the observed value. 
For example, if $q = 0.05$, then the upper quantile is $(1-\frac{q}{2}) = (1 - 0.05/2) = 0.975$ or 97.5\%,  the lower quantile is $\frac{q}{2} = 0.05/2 = 0.025$ or 2.5\%, and the confidence interval is $(1-q) = (1 - 0.05) = 0.95$ or 95\%.

Note that MIS defined in Eq.~\ref{eq_mis} penalizes for two factors: (1) a large interval width $(q_{u_{ij}}-q_{l_{ij}})$, and (2) the observed value being either higher or lower than the upper or lower quantiles. 
MIS  focuses only on calculating intervals by examining the upper and lower quantiles. It does not taken into account the median/ mean prediction. For example, the upper quantile value is 63.33, the lower quantile value is 60.97, and the observed value is 58.87. Then, MIS is (63.33 - 60.97) + 2/0.05 (60.97 - 58.87) = 2.36 + 40* 2.1 = 82.36. If the observed value is 61.87, then the MIS is just the interval (63.33 - 60.97) = 2.36. 


The MIS score measures how narrow a given confidence interval defined by the quantiles is and whether the ground truth is within the interval or not. Additional details on this score can be found in \cite{gneiting2007strictly}. 

The experimental evaluation was conducted on a GPU-based cluster 
with 126 nodes. Each of them contains two 2.4 GHz Intel Haswell E5-2620 v3 CPUs and one NVIDIA Tesla K80 dual-GPU card (two K40 GPUs), enabling two model trainings per node.
The code was implemented in TensorFlow 2.4 and used Python 3.7.0.
We used SDV package (v0.14.0)~\cite{7796926} for fitting the GC models, where one-hot encoding was used for handling the categorical hyperparameter values.

For the hyperparameter search (HPS), we use the learning rate in the continuous range [.0001, 0.02], learning rate decay ratio in the continuous range [0.001, 0.02], filter type in the categorical range [Laplacian, random walk, dual random walk], batch size [8, 16, 32, 64, 128, 256], maximum diffusion steps [1, 2, 3, 4, 5], number of layers for the encoder  [1, 2, 3, 4, 5], number of layers for the decoder [1, 2, 3, 4, 5], number of RNN units per layer  [4, 5, 6,..,128], number of epoch  [20,21,...,100], and  threshold for gradient clipping [1,2,3,...,10] in the discrete range.

\subsection{Impact of scaling hyperparameter search on ensemble accuracy}

Here we show that scaling the HPS results in an increased number of high-performing hyperparameter configurations that can improve the quality of Gaussian copula model and the overall ensemble accuracy.

For the scaling experiments, we ran DeepHyper HPS runs with 10, 25, 50, and 100 workers, each with 
12 hours of wall time as the budget. We note that each compute node in the cluster can run up to 2 workers simultaneously. These runs resulted in 49, 108, 203, and 453 model evaluations, respectively. The default hyperparameter configuration achieved a validation SQR loss of $2.46$. Therefore, we consider the hyperparameter configurations as high  performing when they have a validation loss less than $2.46$.

\begin{figure}[!ht]
\centering
   \includegraphics[width=\linewidth,height=0.5\linewidth]{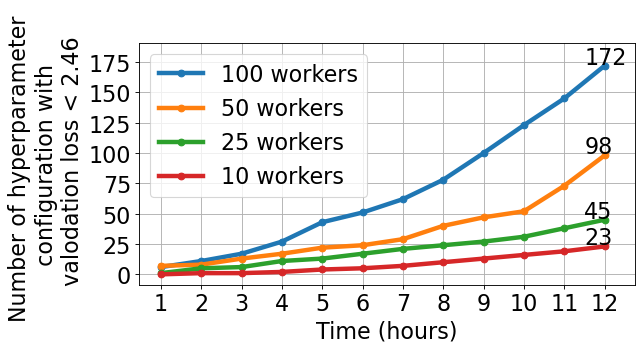}  
\caption{Hyperparameter configurations with validation loss less than 2.46 for different number of workers in DCRNN hyperparameter search. The plot shows that scaling hyperparameter search is critical for finding a large number of high-performing configurations.} 
\label{fig_scaling}
\end{figure}

Figure \ref{fig_scaling} shows the cumulative number of high-performing configurations obtained by DeepHyper as a function of search time. We observe that the number of high-performing configurations increases almost linearly with the increase in the number of workers. In particular, with 100 workers, the search achieves 50 high-performing configurations within 6 hours, whereas the search with 50 workers takes 10 hours to reach that number. In 6 hours, the 100-worker search finds more high-performing configurations than the  25- and 10-worker  searches find. 
The number of high-performing configurations obtained by 10-, 25-, 50-, and 100-worker searches is 23, 45, 98, and 172, respectively.

\begin{figure}[ht]
    \centering
     \includegraphics[width=\linewidth,height=0.5\linewidth]{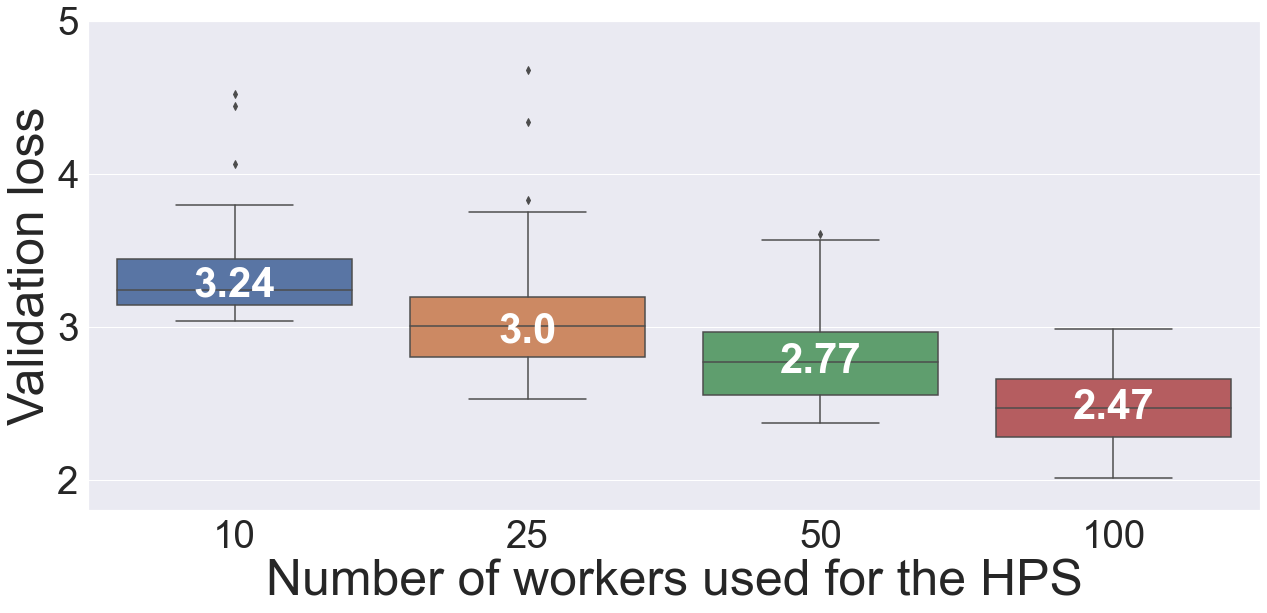}  
\caption{Validation loss distribution of 100 models sampled from a Gaussian copula trained with HPS results with different workers.} 
\label{fig_scaling_val_loss}
\end{figure}

Next we selected a number of high-performing configurations from the DeepHyper HPS runs for generative modeling. For a given worker-count search run, we computed a 10\% quantile of the validation loss values of all hyperparameter configurations. We then selected the configurations that were less than or equal to that cut-off value. These configurations were used to fit a Gaussian copula (GC) model. We applied this procedure for each worker-count search run and denoted the resulting five GC models as GC-10, GC-25, GC-50, and GC-100. Table \ref{tab_GC_time} lists the time taken by each GC model to fit the configurations and sample new hyperparameter configurations from the distribution. We observe that time required for fitting the GC model and sampling from the model is negligible (less than 3 seconds for fitting and 0.04 second for sampling on a single CPU node). 
Once they were fitted, we sampled 100 new hyperparameter configurations from each GC model and trained all them simultaneously on 50 K80 nodes (two model training per node). 

\begin{table}
\centering
\begin{tabular}{|l|l|l|} 
\cline{2-3}
\multicolumn{1}{l|}{} & Fit time (sec) & Sample time (sec)  \\ 
\hline
GC-10~                & 2.57           & 0.030              \\ 
\hline
GC-25                 & 2.82           & 0.031              \\ 
\hline
GC-50                 & 2.89           & 0.032              \\ 
\hline
GC-100                & 2.96           & 0.038              \\
\hline
\end{tabular}
\caption{Time taken to fit the configurations to the GC models (GC-10, GC-25, GC-50, and GC-100) and sample 100 configurations from the trained GC models.}
\label{tab_GC_time}
\end{table}
 
Figure \ref{fig_scaling_val_loss} shows a box plot of the validation loss distribution of 100 trained models from GC-10, GC-25, GC-50, and GC-100. The results clearly show that the validation loss significantly decreases as we move to GC models trained with a larger number of workers. The median values of the validation loss distributions are 3.24, 3.0, 2.77, and 2.47 for GC models trained with 10-, 25-, 50-, and 100-workers HPS results, respectively.

For model selection, we selected the top 25 models from the generated 100 based on the validation loss. The number 25 is motivated by the fact that the state-of-the-art SG-MCMC method used 25 posterior sample models from the trained model. 

\begin{figure}[ht]
  \centering
   \includegraphics[width=\linewidth,height=0.5\linewidth]{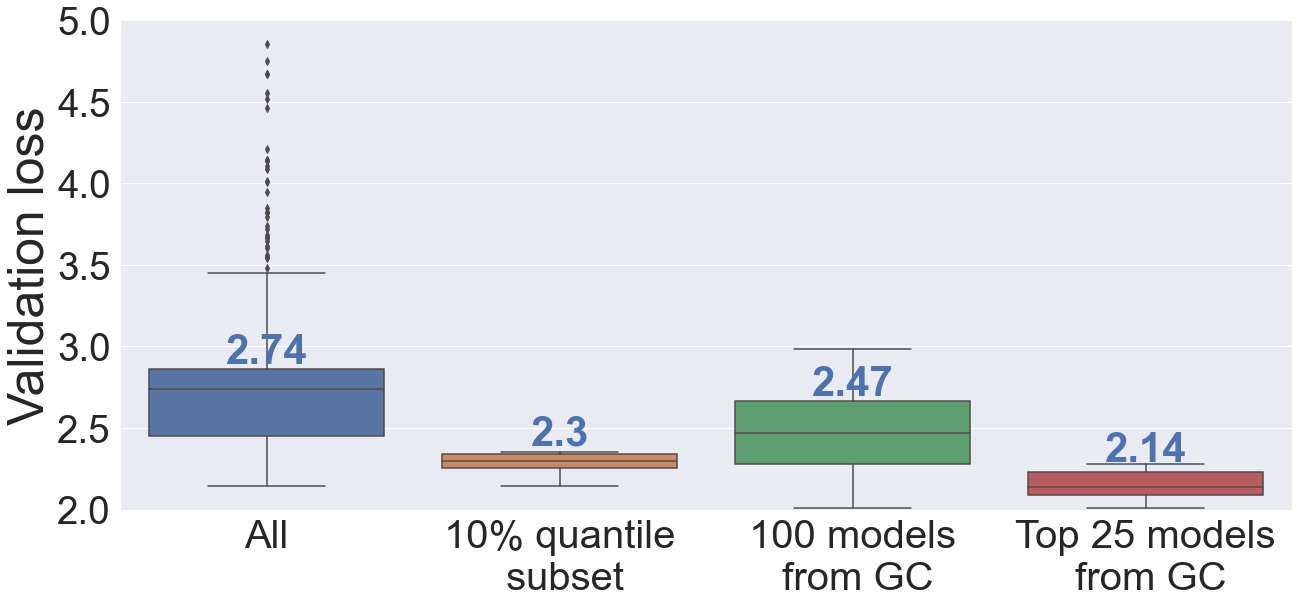}  
  \caption{Validation loss distributions of four settings: (1) all the hyperparameter configurations in the HPS run with 100 workers, (2) 10\% quantile subset of the HPS run used for training the GC-100 model, (3) 100 models from GC-100, and (4) the top 25 of the 100 models sampled from GC-100} 
  \label{fig_gc}
\end{figure}

We also compared the validation loss distributions of (1) all the hyperparameter configurations in the HPS run, (2) 10\% quantile subset of the HPS run used for training the  GC-100 model, (3) 100 models from GC-100, and (4) the top 25 of 100 models from GC-100.
The results are shown in Figure \ref{fig_gc}. 
The results show that the validation loss distribution achieved by the top 25 models is superior to the hyperparameter values found by the search alone. Moreover, using all 100 models from the GC model results in a validation loss distribution that is better than that of HPS run. The strategy of generating models from GC and selecting a subset from them seems a good strategy. We expect that by generating a large number of high-performing configurations by scaling HPS further and/or by using faster GPUs, we can train better GC models, which will result in overall improvement in the ensemble validation loss distribution.  

\begin{figure}[ht]
    \centering
     \includegraphics[width=\linewidth,height=0.5\linewidth]{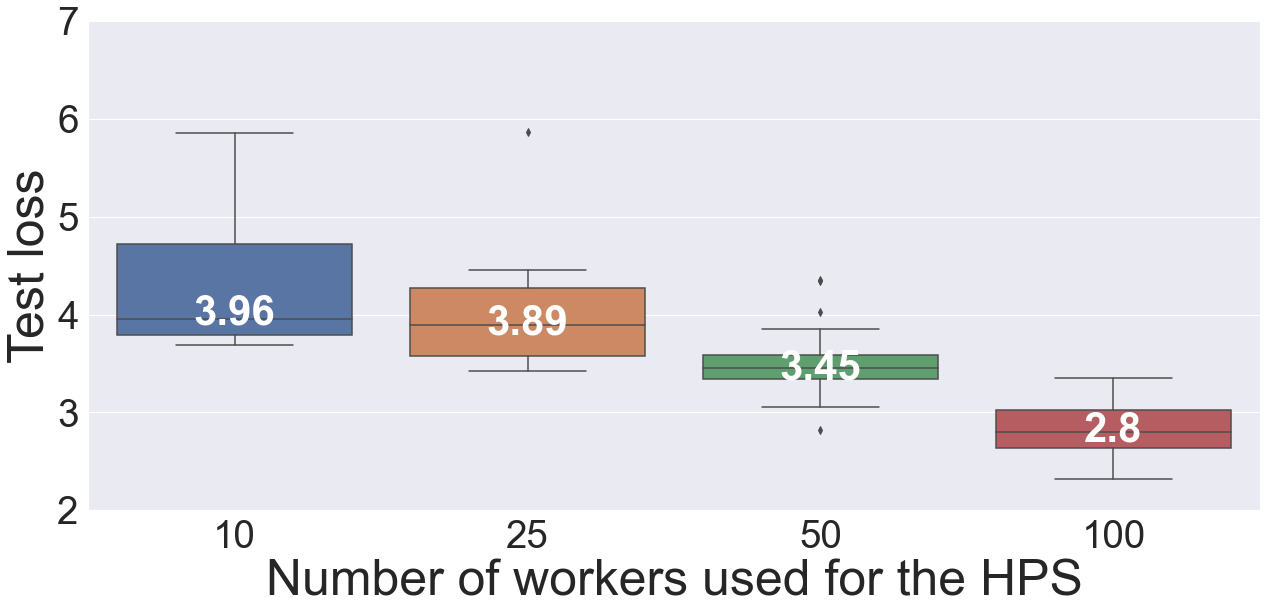}  
\caption{Test loss distribution of top 25 out of 100 models sampled from Gaussian copula trained with results from HPS with different workers.} 
\label{fig_scaling_test_loss}
\end{figure}

Using the selected 25 models, we performed  inference on the test data and computed the test loss. The results are shown in Figure \ref{fig_scaling_test_loss}. We observe that the 25 models generated with a larger number of workers for the HPS significantly reduce the test loss. The trend is similar to the validation loss distribution with 100 models. The medians of the distribution are 3.96, 3.89, 3.45, and 2.8 for 10, 25, 50, and 100 workers, respectively.  The findings lead us to conclude that when conducted on a large number of workers, HPS yields better configurations, which result in improved overall predictive accuracy of the ensembles.


\subsection{Uncertainty estimation and comparison with the state-of-the-art}
\begin{table*}[t]
\begin{scriptsize}
\centering
\begin{tabular}{|l|l|l|l|l|l|l|l|l|l|l|l|}
\hline
T                        & Metric       & Point & Bootstrap  & Quantile & SQ  & MIS & MC Dropout  & SG-MCMC & {\bf DCRNN-SQRUQ}  \\ \hline
\multirow{4}{*}{15 mins} & MAE (mean)   & 2.38  & 2.63          & 2.43         & 2.67   & 2.63   & 2.47           & 2.32        & {\bf 2.21} \\ \cline{2-10} 
                         & MAE (median) &       &               &              &        &        &                &             & {\bf 2.20}         \\ \cline{2-10} 
                         & MIS          &       & 39.76         & {\bf 18.32}       & 29.04  & 18.26  & 27.61          & 32.21       & 24.79      \\ \cline{2-10} 
                         & Interval     &       & {\bf 5.8}           & 12.42        & 8.38   & 12.46  & 8.99           & 8.73        & 8.29           \\ \hline
\multirow{4}{*}{30 mins} & MAE (mean)   & 2.73  & 3.19          & 2.79         & 3.1    & 3.08   & 2.94           & 2.54        & {\bf 2.44}\\ \cline{2-10} 
                         & MAE (median) &       &               &              &        &        &                &             & {\bf 2.45}       \\ \cline{2-10} 
                         & MIS          &       & 38.48         & 21.54        & 40.93  & {\bf 21.09}  & 33.38          & 31.87       & 27.4       \\ \cline{2-10} 
                         & Interval     &       & {\bf 7.86}         & 13.48        & 8.37   & 13.99  & 11.1           & 12.62       & 10.53       \\ \hline
\multirow{4}{*}{60 mins} & MAE (mean)   & 3.14  & 3.99          & 3.19         & 3.75   & 3.65   & 3.7            & 3           & {\bf 2.83} \\ \cline{2-10} 
                         & MAE (median) &       &               &              &        &        &                &             & {\bf 2.80}    \\ \cline{2-10} 
                         & MIS          &       & 38.58         & 25.74        & 60.56  & {\bf 24.33}  & 43.11          & 30.35       & 29.33     \\ \cline{2-10} 
                         & Interval     &       & 11.65         & 14.5         & {\bf 8.38}   & 15.55  & 14.46          & 18.79       & 12.34        \\ \hline
\end{tabular}
\caption{Performance comparison of different UQ methods}
\label{tab_comp}
\end{scriptsize} 
\end{table*}

In this section we compare the uncertainty estimates from our proposed DCRNN-SQRUQ with several other DCRNN UQ techniques that were used in a recent study \cite{wu2021quantifying} and show that our method outperforms all other methods. For the experimental comparison, we adopt the same setup in terms of error metrics, forecasting horizons, and confidence interval as described in \cite{wu2021quantifying}.



The uncertainty estimation techniques reported in \cite{wu2021quantifying} are as follows.
(1) Bootstrap \cite{efron2016computer}: This method randomly samples 50\% of training data points to train DCRNN  model. Prediction is performed on the entire test dataset, and 25 samples are collected in order to calculate mean predictions and confidence intervals.
(2) Quantile \cite{koenker2005quantile}: This method uses pinball loss for DCRNN, and the prediction is performed for three different quantiles 2.5\%, 50\%, and 97.5\% (95\% confidence interval).
(3) Spline quantile regression (SQ) \cite{gasthaus2019probabilistic}: This method models the quantile functions using linear splines and optimizes the continuous ranked probability score as a loss function. 
(4) Mean Interval Score  \cite{wu2021quantifying}: MIS is directly used in the loss function along with MAE, and a multiheaded model is used to jointly output the upper bound, lower bound, and prediction for a given input. 
(5) Monte Carlo  Dropout \cite{gal2016dropout}: This method uses the algorithm develops by Zhu et al. \cite{zhu2017deep}. A 5\% random drop rate is used during the testing time over 50 iterations to get a stable prediction.
(6) Stochastic Gradient Markov Chain Monte Carlo (SG-MCMC) \cite{ma2015complete}: This is the current state of the art Bayesian neural network for DCRNN and uses a subsampling technique to minimize the cost of Markov chain Monte Carlo (MCMC) per iteration. A Gaussian prior $\mathcal{N}$(0, 0.4) with randomly initialization of $\mathcal{N}$(0, 0.2) is used for the model parameters. 
The mean and confidence intervals are obtained by averaging from 25 posterior model samples.

The comparison results are shown in the Table~\ref{tab_comp}. The results of the six methods (Bootstrap, Quantile, SQ, MIS, MC Dropout, SG-MCMC) are taken directly from \cite{wu2021quantifying}. The performance of each method is evaluated by using the error metrics MAE and MIS (95\% confidence interval). The comparison is done on three different forecasting horizons: 15 minutes, 30 minutes, and 60 minutes. 


The results show that our DCRNN-SQRUQ method outperforms the state-of-the-art SG-MCMC method and a number of other UQ methods. The mean and the median MAE of our DCRNN-SQRUQ shows a better prediction accuracy in MAE than does SG-MCMC across all three forecasting horizon. Moreover, the MIS score and interval (defined in Eqs. \ref{eq_mis} and \ref{eq_interval})  
show that our method achieves a low MIS score and interval with a narrow interval compared with other techniques. An exception is for the MIS method that directly minimizes the MIS score in the loss function;  as expected, it achieves the best MIS performance. However, the prediction accuracy and interval score of DCRNN-SQRUQ are better than those found by MIS. 
Overall, our proposed DCRNN-SQRUQ is more robust in prediction accuracy and achieves MIS and interval scores that are better than the state of the art SG-MCMC method.

\subsection{Uncertainty vs traffic dynamics}
Here we estimate aleatoric and epistemic uncertainty using DCRNN-SQRUQ and investigate the relationships between the traffic dynamics and the estimated uncertainty measures. According to this analysis, increased traffic dynamics produced by fast changes in traffic behavior result in high aleatoric and epistemic uncertainty. Currently, none of the existing UQ methods developed for DCRNN provide separate aleatoric and epistemic uncertainties.

Traffic dynamics describe the change in traffic behavior observed by the variability of the traffic speed. High dynamics indicate instability in the speed caused by rapid change of traffic behavior, whereas low dynamics indicate stable behavior of the traffic. 
The traffic dynamics at a given sensor location can be measured by the coefficient of variation (COV) \cite{mallick2020graph} of the time series data (speed). The COV becomes large (small) when the variations/change in the speed values are high (low). To measure COV, we calculated the mean ($\mu$)  and standard deviation ($\sigma$) of the speed of each sensor location over the entire timeline. The COV is estimated by $\sigma$/$\mu$. Afterwards, the COV on METR-LA dataset are binned in four ranges: (1) $0.2 <$ COV $< 0.3$;  (2) $0.3 <$ COV $< 0.4$; (3) $0.4 <$ COV $< 0.5$; and (4) $0.5 <$ COV. The binning helps us identify the sensor location with low to high traffic dynamics. Of a total of 207 sensor locations, 34 locations have COV between 0.2 and 0.3,  120 locations have  COV between 0.3 and 0.4,  42 locations have COV between 0.4 and 0.5, and 11 locations have COV more than 0.5.

\begin{figure}[ht]
  \centering
   \includegraphics[width=\linewidth,height=0.5\linewidth]{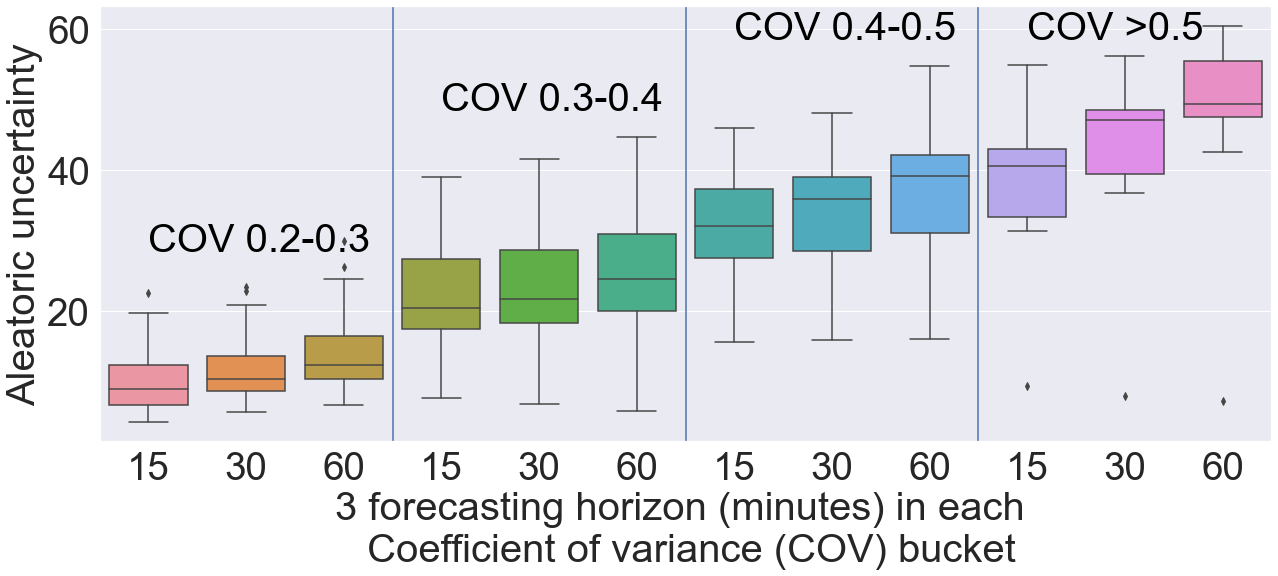}  
\caption{Coefficient of variation of the sensors  binned into four categories. Box plots depict the aleatoric uncertainty estimation for each bin on the 15-, 30-, and 60-minute forecasting horizons using DCRNN-SQRUQ.}
\label{fig_cov_ale}
\end{figure}

\begin{figure}[ht]
  \centering
   \includegraphics[width=\linewidth,height=0.5\linewidth]{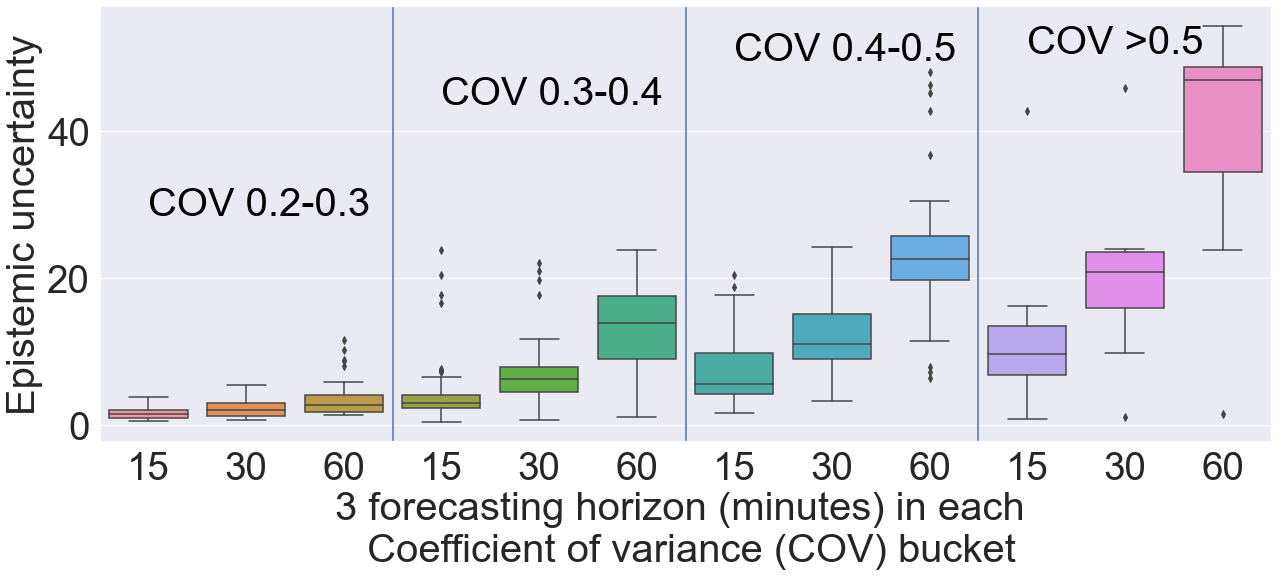}  
\caption{Coefficient of variation of the sensors  binned into four categories. Box plots depict the epistemic uncertainty estimation for each bin on the 15-, 30-, and 60-minute forecasting horizons using DCRNN-SQRUQ.}
\label{fig_cov_epi}
\end{figure}

We estimate the aleatoric and epistemic uncertainty using  Eq.~\ref{eq:var-decomposition-2}. The aleatoric uncertainty is the mean of the prediction variance of the 2.5\% and 97.5\% quantile over the top 25 models selected from the ensemble. 
For each sensor location we now have the COV value and the aleatoric uncertainty measure.  We separate the sensor locations based on the COV values into the four bins mentioned above and plot the distribution of aleatoric uncertainty across three forecasting horizons in box-and-whisker plot in Figure \ref{fig_cov_ale}.  We  have three forecasting horizons and four COV bins, for a total of 12 box plots per figure.
Across all the COV bins we can observe that 25\%, median, and 75\% quantiles of the aleatoric uncertainty increase with the increasing COV values. The box plots for the 15-, 30-, and 60-minute forecasting horizons show a similar trend. This result indicates that the aleatoric uncertainty is high on the sensor locations with high traffic dynamics or COV values. Since the aleatoric  uncertainty is irreducible,  we cannot mitigate it. In a proactive traffic management system, however, aleatoric uncertainty can be used to estimate  worst-case traffic management and planning.

\begin{figure}[!t]
\centering
\begin{subfigure}{.5\textwidth}
  \centering
  \includegraphics[width=\linewidth,height=0.5\linewidth]{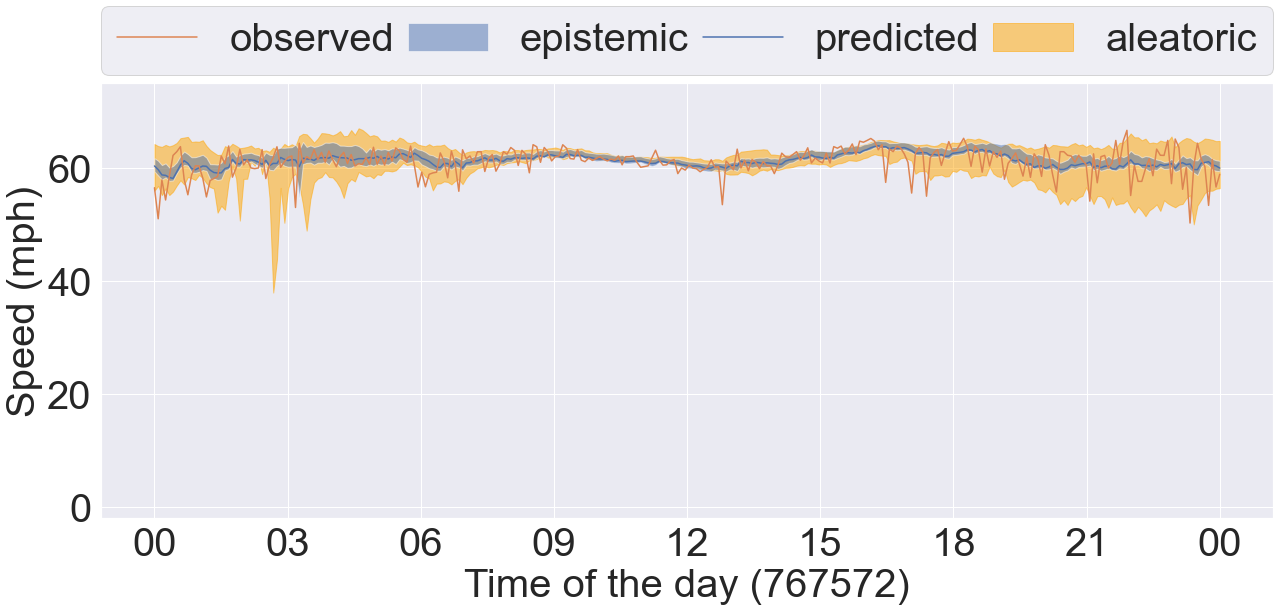}  
  \caption{COV between 0.2 to 0.3}
  \label{fig_plot1}
\end{subfigure}
\begin{subfigure}{.5\textwidth}
  \centering
  \includegraphics[width=\linewidth,height=0.5\linewidth]{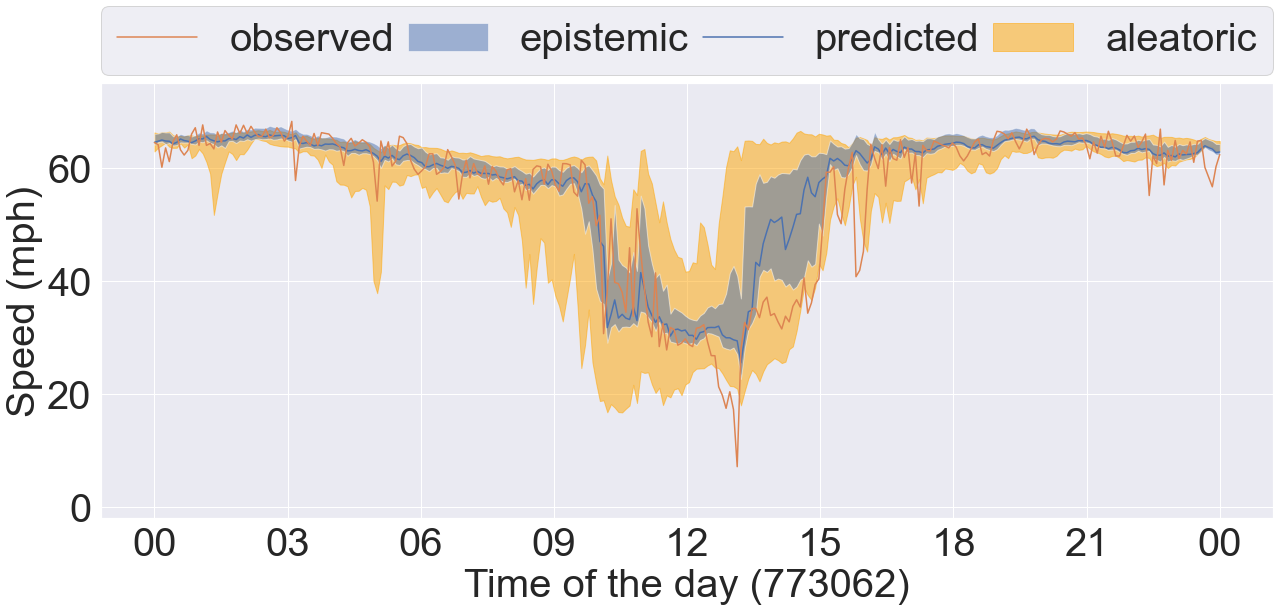}  
  \caption{COV between 0.3 to 0.4}
  \label{fig_plot2}
\end{subfigure}
\begin{subfigure}{.5\textwidth}
  \centering
  \includegraphics[width=\linewidth,height=0.5\linewidth]{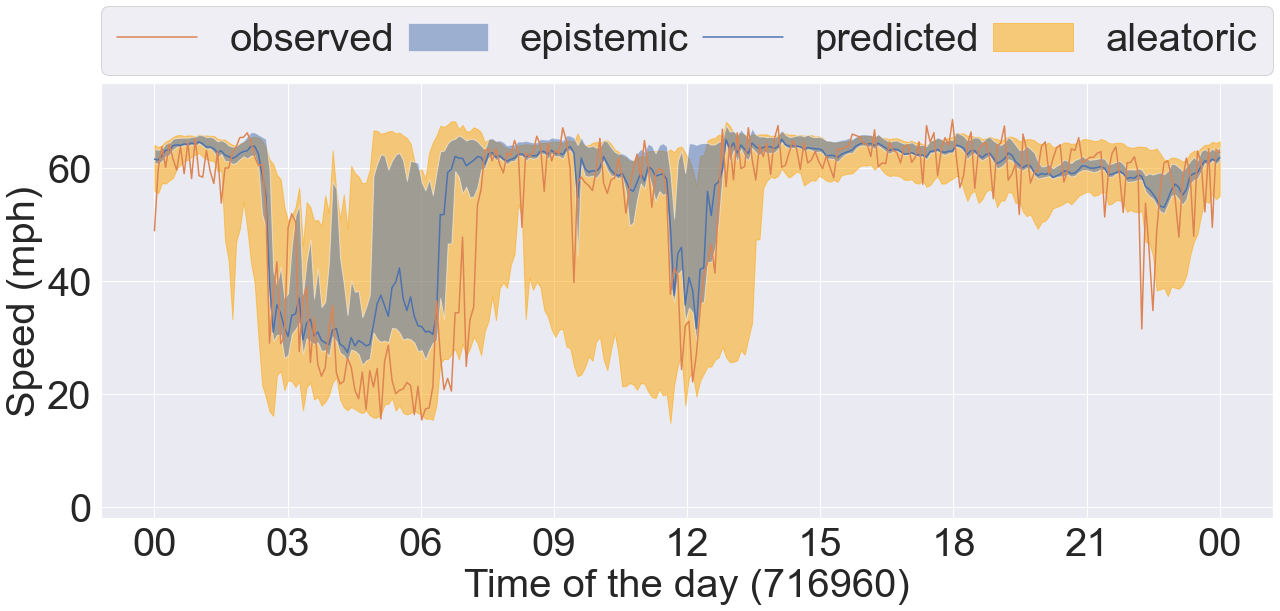}  
  \caption{COV between 0.4 to 0.5}
  \label{fig_plot3}
\end{subfigure}
\begin{subfigure}{.5\textwidth}
  \centering
  \includegraphics[width=\linewidth,height=0.5\linewidth]{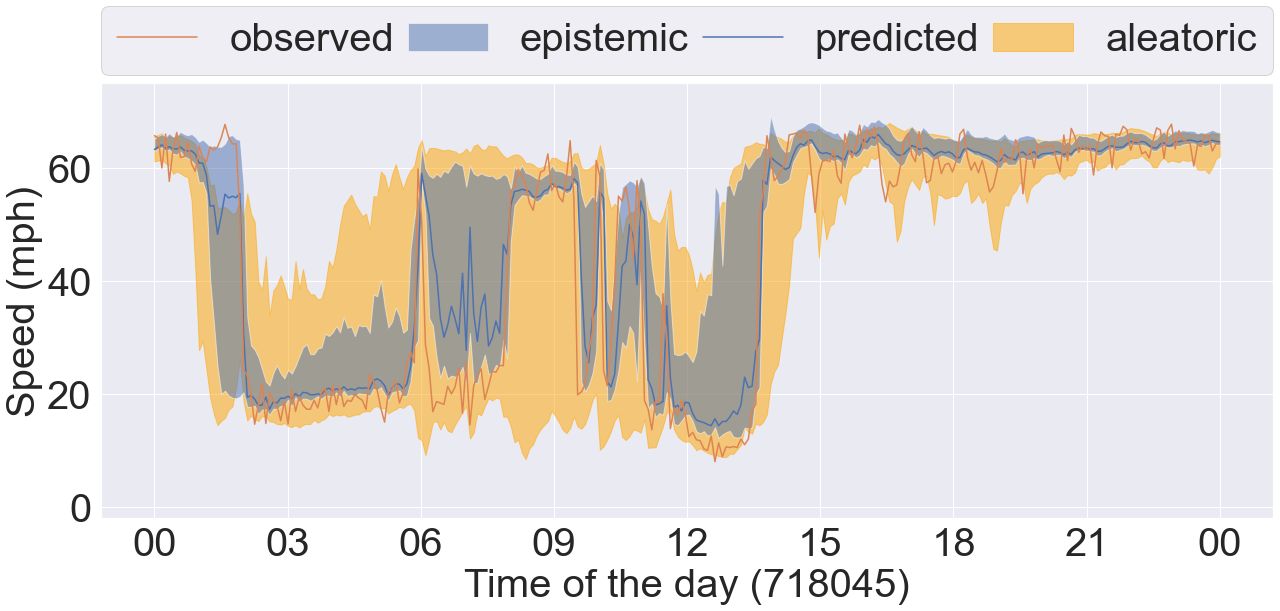}  
  \caption{COV above 0.5}
  \label{fig_plot4}
\end{subfigure}
\caption{60-min-ahead traffic forecasting on four sensors selected from 4 COV bins. The aleatoric uncertainty or 95\% prediction interval is shown in yellow shade, and the epistemic uncertainty is shown in blue shade.}
\label{fig_plot}
\end{figure}

Next, we estimate the epistemic uncertainty by using Eq. \ref{eq:var-decomposition-2}. Epistemic uncertainty measures the variance of the mean across top 25 models selected from the ensemble. Similar to the aleatoric uncertainty, we separate the sensor locations based on COV values into  four bins and plot the distribution of epistemic uncertainty across three forecasting horizons in box-and-whisker plots in Figure \ref{fig_cov_epi}. A similar trend can be observed here: the epistemic or  model uncertainty increases for the sensor locations having high COV value or traffic dynamics. 
The epistemic uncertainty for 60-minute forecasting is high for COV $> 0.5$ (9th box in the Figure \ref{fig_cov_epi}). The predictive uncertainty of the model is high due to the high traffic dynamics. Therefore, we recommend performing 15-minute forecasts for those sensor location where COV $> 0.5$ (11 sensors belong to this bin). The 15-minute forecast (7th box in Figure \ref{fig_cov_epi}) has a median of the distribution of the epistemic uncertainty below 10. Similarly, uncertainty is high for 60-minute forecasting in the COV bin between 0.4 to 0.5. Hence, we recommend  performing 30-minute forecasting for those sensor locations (42 sensors belong to this bin). The ability to provide these uncertainty measures and consequent recommendations is crucial to putting these models into practice. Limiting the forecasting horizon for highly dynamic sensors will stabilize the DCRNN prediction and minimize the risk of using the forecasting results for crucial decision making.

\begin{figure}[!t]
\centering
   \includegraphics[width=\linewidth,height=0.5\linewidth]{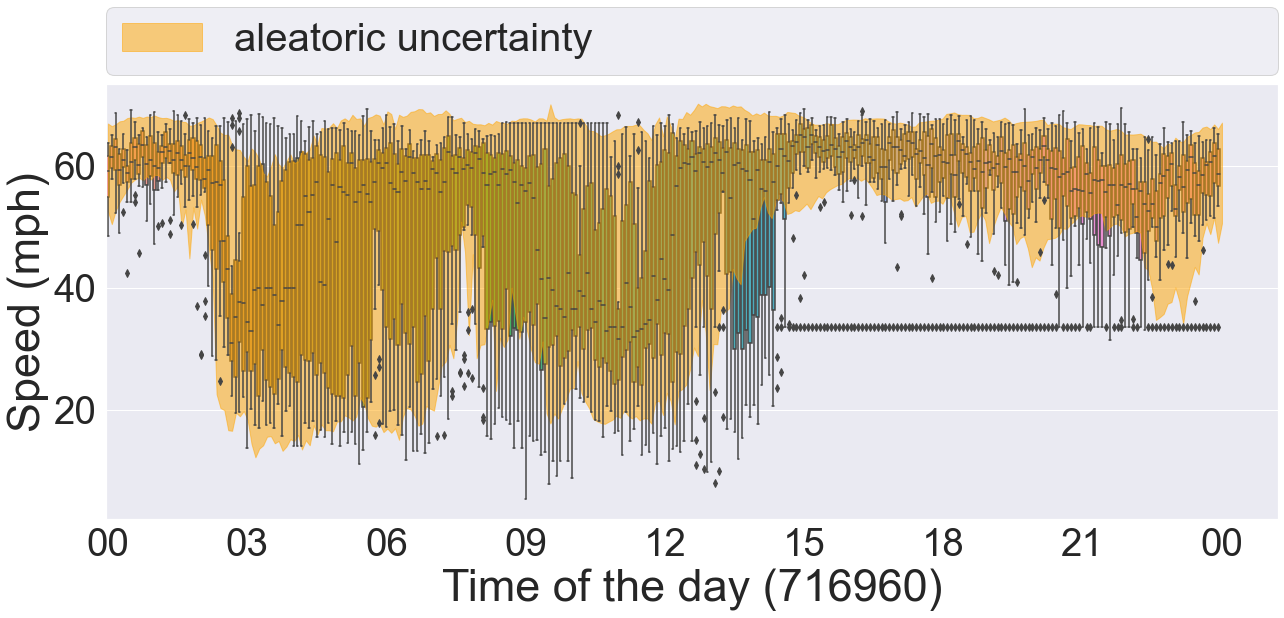}  
\caption{Time series data for sensor ID: 716960 over multiple days demonstrating the variability in the data. The aleatoric uncertainty on top of the daily variation indicates how well DCRNN-SQRUQ captures the variability.}
\label{fig_high_ale}
\end{figure}

Further, we plot 4 sets of time series data taken from 4 different COV bins in Figure \ref{fig_plot}. Each plot contains the observed and predicted timeseries data, and the epistemic and aleatoric uncertainty intervals estimated by our proposed method DCRNN-SQRUQ methods. The results shown in the Figure \ref{fig_plot} are for the 60-minute forecasting horizon. 
The epistemic uncertainty interval is calculated by using the 2.5\% and 97.5\% quantiles of the median prediction of the top 25 models - shown in blue. 
The aleatoric uncertainty intervals are the medians of the 2.5\% and 97.5\% quantile of the top 25 models- shown in yellow. 
Figure \ref{fig_plot1} shows the sensor location with the lowest COV value (between 0.2 and 0.3). Traffic behavior does not change rapidly in this COV bin. Hence, the interval band is narrow for both of the uncertainty measures. Figure \ref{fig_plot2} shows the sensor location with COV value between 0.3 and 0.4. Traffic dynamics are higher compared with the previous bin. 
Both the aleatoric and epistemic uncertainty measures are high around 9 am and 3 pm. The traffic dynamic increases more in Figure \ref{fig_plot3}, and the uncertainty band becomes even wider here.  Epistemic uncertainty estimates are high during the  drop/rise in traffic speed. 
In contrast, aleatoric uncertainty is high around 6  am, 9 am, 12 pm, 
and 11 pm. Specifically, during 9 am the uncertainty is high but the traffic is  stable. This uncertainty is coming from the data. Therefore, to validate it, we plot the timeseries data of the same sensor (sensor ID: 716960) over multiple days in box-and-whisker plot shown in the Figure \ref{fig_high_ale}. The results show huge variation in the speed profile during 3 am, 9 am, 12 pm, and 11 pm. We also plot the aleatoric uncertainty on top of the daily variation of the data. We see that our uncertainty measure covers the variation in the data. This explains the high uncertainty at 9 am in  Figure \ref{fig_plot3}. Likewise, the traffic dynamics increase more in  Figure \ref{fig_plot4} where the COV is greater than 0.5. The aleatoric uncertainty is high from 3 am to 7 am because of the variety of patterns  in the data. The epistemic uncertainty is high  during the drop/rise of traffic speed, which is difficult for the model to capture.

\section{Related work}


Quantifying predictive uncertainty is important for deep learning models. 
Various methods for quantifying uncertainty, such as \cite{abdar2021review, hullermeier2021aleatoric}, have been investigated over the past decade.  In this paper  we look at the uncertainty quantification techniques that are used specifically in traffic forecasting.

Uncertainty quantification has not yet been  extensively adopted for traffic forecasting.
Only a few attempts have been made to measure uncertainty in highway traffic forecasts.
The generalized autoregressive conditional heteroscedasticity (GARCH) \cite{guo2010real, guo2014adaptive, tsekeris2010short} model was used to estimate the uncertainty of traffic forecasting with a Kalman filter.  A bootstrapping technique ass used by Matas et al. \cite{matas2012traffic} for uncertainty estimation. Quantile regression recently has been  developed to forecast traffic volumes with confidence intervals \cite{hoque2021estimating}. Bayesian neural networks were used \cite{van2009bayesian, mazloumi2011prediction, zhu2017deep} to predict the travel time along with the confidence intervals.  All of these methods, however, perform forecasting on individual locations. None of the  methods consider spatiotemporal dynamics of all the locations together in their models.

Recently, Wu et al.~\cite{wu2021quantifying} developed and tested six uncertainty estimation techniques for DCRNN. These include five frequentist methods---bootstrap,  MIS,  quantile regression, spline quantile regression,  Monte Carlo  dropout, and one Bayesian model trained through 
SG-MCMC. 
They compared these strategies empirically and reported on the effectiveness and computational trade-offs of various UQ methods. According to their experimental results, Bayesian DCRNN with SG-MCMC results in superior mean prediction when compared to the five frequentist methods but it is computationally expensive. Frequentist approaches, on the other hand, are inexpensive and effective in covering data variations. The uncertainty calculated in Bayesian DCRNN is the total uncertainty, a combination of both epistemic and aleatoric; it cannot distinguish between the two types of uncertainty. Moreover, Bayesian neural networks are computationally more expensive to train than their deterministic counter parts and typically do not scale well for large data sets due to the sequential nature of the specialized training procedures.

Our method is robust in mean predictions and is effective in covering variations in the data. Moreover, the hyperparameter estimation and ensemble can be scaled over multiple compute nodes. Overall, our method is more effective and more efficient compared with other Bayesian and frequentist approaches in terms of accuracy and scalability.

Our approach was inspired by four recent works on uncertainty estimation through deep ensembles: deep ensemble~\cite{lakshminarayanan2017simple}, hyper ensemble~\cite{wenzel2020hyperparameter}, neural ensemble search \cite{zaidi2021neural}, and joint neural and hyperparameter search ensembles \cite{egele2021autodeuq}. 
These methods assume a Gaussian distribution for the output variable, perform regression on a single variable, do not make use of scale for hyperparameter search to generate ensembles, and do not adopt generative modeling. The uniqueness of our method stems from 1) its application to spatial temporal graph neural networks, 2) inclusion of Gaussian-assumption-free simultaneous quantile regression, 3) multi-output spatial temporal regression, 4) the use of hyperparameter search at scale, and 5) Gaussian copula generative modeling for ensemble construction.

\section{Summary and conclusion}
We developed DCRNN-SQRUQ, a scalable Gaussian assumption-free ensemble-based uncertainty quantification for traffic forecasting for spatial temporal graph neural networks. DCRNN-SQRUQ it employs simultaneous quantile regression loss for DCRNN, Bayesian hyperparameter search, generative modeling with a Gaussian copula mode, ensemble training, and variance decomposition for Gaussian quantile regression ensembles. 
DCRNN with SQR effectively captures  well-calibrated forecasting intervals. The Bayesian optimization-based scalable hyperparameter search efficiently finds the most promising hyperparameter configuration. We also showed that scaling the hyperparameter search is important for finding promising hyperparameter configurations and for improving the forecasting accuracy of the ensembles. The Gaussian copula generative model is inexpensive yet efficient in generating more high-performing hyperparameter configurations from promising hyperparameter configurations found in a hyperparameter search. Furthermore, we developed a variance decomposition methodology for DCRNN ensembles with SQR loss function and use it for estimating the aleatoric (data) uncertainty and epistemic (model) uncertainty.

We demonstrated that our approach outperformed the state-of-the-art Bayesian neural network method and a variety of other frequentist uncertainty techniques.
Experiments on a real-world dataset showed that our method is robust in mean predictions and effective in capturing data variations. 
In addition, we investigated the relationship between aleatoric/epistemic uncertainty and traffic dynamics. The analysis showed that increased traffic dynamics caused by rapid changes in traffic behavior resulted in high aleatoric and epistemic levels. Because aleatoric uncertainty is irreducible, it can be used for worst-case traffic planning in proactive traffic management. The epistemic uncertainty can be used to identify the limitations of the model and appropriate location-specific forecasting horizons. Currently, no other UQ method can provide separate aleatoric and epistemic uncertainty estimates for the DCRNN and, in general, for any other traffic forecasting methods. From a practitioner perspective, it provides a mechanism for reducing the forecasting horizon when the uncertainty measure is too high. This is very important for reducing the risk of using the model in real-world decision making.

The key advantages of our approach are generality and scalability. Our proposed approach for building model ensembles is based on a scalable hyperparameter search and computationally inexpensive generative modeling. In principle, we can apply our UQ approach to any neural network models. This involves taking the default neural network architecture, using simultaneous quantile regression loss, running hyperparameter search, training a generative model from the hyperparameter search runs, performing model selection, and decomposing the uncertainty estimates. The scalability of the approach is derived from scalable hyperparameter search and scalable ensemble training. Both phases provide improvement with respect to ensemble predictive accuracy and uncertainty estimation. 

Our future work will include (1) application of the proposed UQ methods for various spatial and temporal forecasting neural network models; (2) comparison with other forms of Bayesian neural networks; (3) advanced generative modeling with autoencoders; and (4) uncertainty-informed traffic planning using reinforcement learning.

\clearpage


\bibliographystyle{vancouver} 
\bibliography{uq}

\end{document}